\documentclass[11pt]{article}

\usepackage[final]{acl}

\usepackage{times}
\usepackage{latexsym}
\usepackage{amsmath,amssymb,amsfonts}
\usepackage{algorithmic}
\usepackage{subcaption}
\usepackage{textcomp}
\usepackage{xcolor}
\usepackage{multirow}
\usepackage{booktabs}
\usepackage{tabularx}
\usepackage{array}
\usepackage{url}

\usepackage[T1]{fontenc}
\usepackage[utf8]{inputenc}

\usepackage{microtype}

\usepackage{inconsolata}

\usepackage{graphicx}

\title{Visual Input and Its Framing Undermine the Stability of Attribute-based Descriptions Produced by Large Vision-Language Models}

\author{
Xiaomeng Wang\textsuperscript{1} \quad
Martha Larson\textsuperscript{1} \quad
Ping Mei Vogels\textsuperscript{1} \quad
Zhengyu Zhao\textsuperscript{2} \\
\textsuperscript{1}Radboud University
\qquad
\textsuperscript{2}Xi'an Jiaotong University \\
\texttt{xiaomeng.wang@ru.nl} \quad
\texttt{m.larson@cs.ru.nl} \quad
\texttt{pingmei.vogels@ru.nl} \\
\texttt{zhengyu.zhao@xjtu.edu.cn}
}

\begin{document}
\maketitle

\begin{abstract}
Large vision-language models (LVLMs) are commonly prompted with text alone or together with an image.
In this paper, we find a new aspect of LVLM vulnerability in its response robustness: adding an image can unexpectedly affect responses even when \textit{text prompts ask about a concept rather than the instance of the concept depicted in the image}.
For example, adding an image of an animal can change the generated attribute descriptions of its breed, and framing of the image further affects those changes.
We study this new problem on seven LVLMs with image sets
covering 30 animal breeds and eight food categories, using two types of framings that focus on the subject (subject-focused) or place it in a broader context (subject-in-situation).
We draw the following main findings:
(1) adding images increases physical-term proportions for animals (16.7\% $\to$ 32.8\%) and decreases sensory-term proportions for food (83.6\% $\to$ 73.2\%) among the most frequent terms, averaged over seven models and two framings on correctly recognized images.
(2) attribute-term distributions differ significantly between the two framing types for 41.4\% of animal and 71.4\% of food model--concept pairs, with generally higher physical-term proportions for animals and sensory-term proportions for food under subject-focused than subject-in-situation framing.

\end{abstract}

\section{Introduction}
\label{sec:intro}
%ML: Function: Introduce the importance of stability in the case of prompts that require stability
Large vision-language models (LVLMs) can process text prompts alone or together with images.
LVLMs' capability to interpret visual input supports systems such as image captioning, assistive technologies for people with visual impairments, and perception modules for autonomous agents~\citep{blip2,zhou2024autonomous,jiang2025benchhuman,karamolegkou2025evaluatingvisually}.
LVLMs in these systems need to provide stable and consistent responses that meet users' needs.
Particularly, when asking about the general properties of a concept, users may name it directly in a text prompt or present an instance of the concept as an accompanying image. 
If the concept is correctly recognized, generated descriptions of the concept should remain consistent regardless of whether and how the visual input is included in the prompt.

\begin{figure}[t]
    \centering
    \includegraphics[width=\columnwidth]{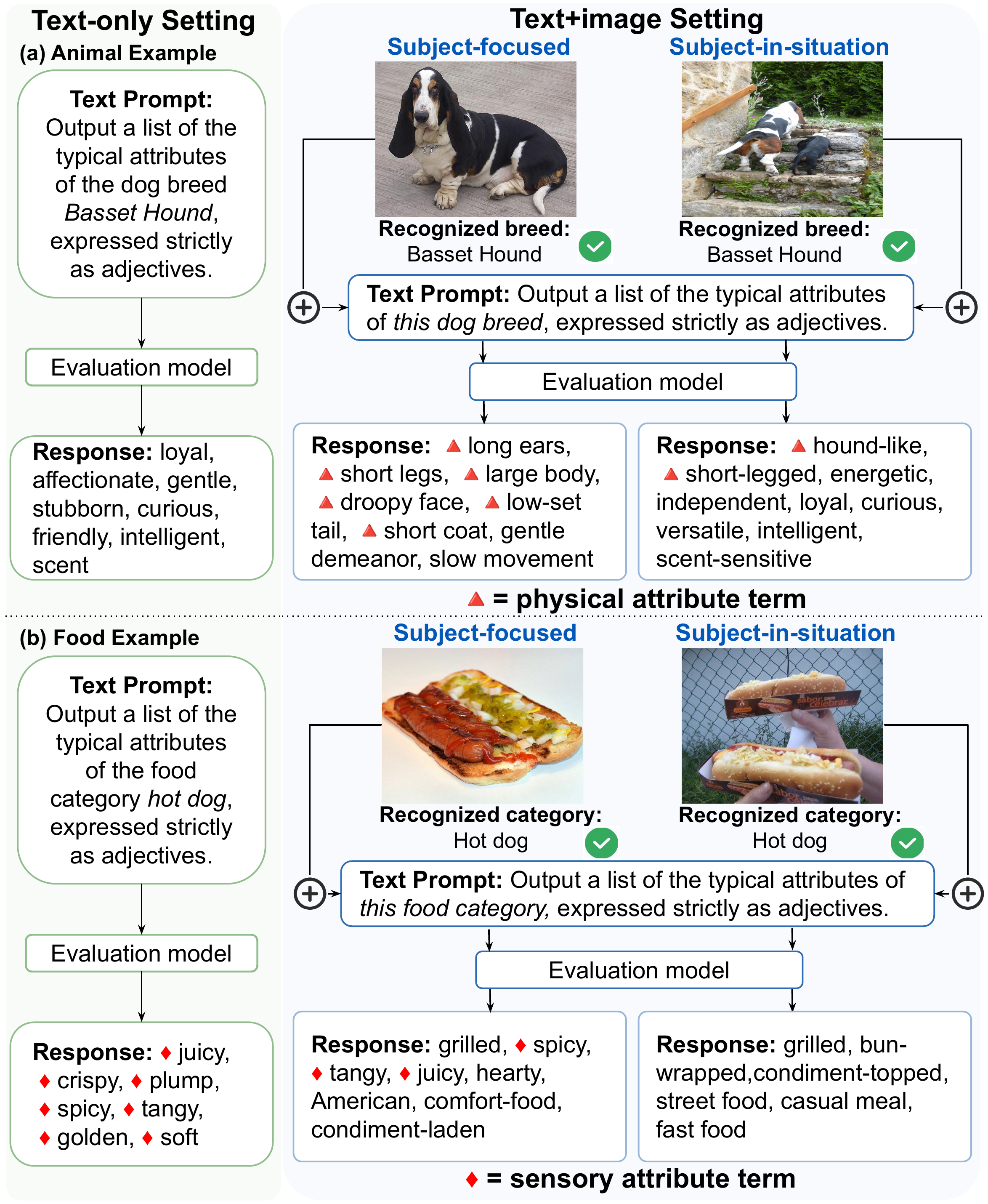}
    \caption{
     When concept-relevant prompts ask about the same concept, responses in the text+image setting shift towards more physical attribute terms for the animal example and fewer sensory attribute terms for the food example compared with the text-only setting. How the depicted subject is framed (subject-focused or subject-in-situation) further affects these shifts.
\label{fig:illustration}}
%zz at a first glance, I didn't get what the triangle and diamond mean. We need to make it more obvious.
%zz consider move the food example to the appendix (only mentioned in the caption is enough because the two examples follow exactly the same logic) and make the rest larger.
\end{figure}

Specifically, when the visual input is included, generated descriptions of the concept must remain stable across changes in \emph{visual framing}, defined as how the depicted subject is visually presented in an image through choices regarding viewpoint, scene, angle, cropping, editing, and image selection~\citep[p.~237]{coleman2010framing}.
Here, the depicted subject is a specific entity, whereas the target concept is the category of which this entity is an instance.
Although framing matters for image-relevant prompts that request image-specific descriptions, it should not affect concept-relevant prompts, which request general properties of the target concept.
For example, descriptions of the general properties should not depend on the depicted dog's pose or activity.
% We use \textit{concept-relevant prompts} to request general attributes of the target concept rather than properties specific to the depicted instance.
% In contrast, \textit{image-relevant prompts} request attributes of the particular image.

%ML: Function: What we do in this paper. Description of the expected behavior. 
%ML Note: Somewhere towards the beginning we need to introduce the notion of "framing" (which is in the title) and why it is important. LVLMs should be sensitive to framing, but only in cases where the framing matters to the prompt. 
%Maybe this is helpful about framing: "Framing is important to how people interpret images. It is related to intent. However, people can also ignore the framing of the image and focus on the main concept it depicts when this is necessary for image interpretation. In this paper, we show that LVLMs struggle to provide stable answers to prompts in the case that they are prompted in a way that requires them to ignore framing."
%ML: Define what "concept-level" description is. What is the connection between a concept and a subject for an image. 
In this paper, we study whether LVLM-generated descriptions of a target concept under concept-relevant prompts remain stable when images are added or their framing changes.
We focus on attribute-based descriptions because they provide a controlled way to examine the semantic properties that a model associates with a concept, while avoiding additional variability introduced by open-ended captioning.
As illustrated in Figure~\ref{fig:illustration}, we compare a text-only setting, where the target concept is provided only as text, with a text+image setting, where the same concept is provided by way of an image.
To control for recognition errors in the text+image setting, we focus on images for which the LVLMs are able to correctly recognize the target concept in the image.
Within the text+image setting, we further compare two visual framing types, motivated by the different roles of object and contextual information in intent recognition~\citep{jia2021intentonomy}.
Subject-focused images present the subject as the dominant visual focus, whereas subject-in-situation images present the subject within a broader activity or environmental context.
% This framing contrast concerns the role of context rather than simply its presence or absence and is not intended as an exhaustive taxonomy of framing.

With a robust LVLM, the attribute-based descriptions of the target concept under a concept-relevant prompt should remain stable regardless of the input conditions.
% When prompting an LVLM to describe attributes of the concept, the target concept can be presented in different ways: it can be given as a textual concept name, or it can be presented via a visual instance of the concept in different framing types.
% Ideally, the response of the LVLM should be the same in both cases, i.e., it should be insensitive to the use of an image and also to visual framing.
% Moreover, once the target concept in an image is correctly recognized, the generated attribute descriptions of the target concept should reflect the concept itself no other image-specific visual cues such as pose, viewpoint, activity, or surrounding context.
% how the subject of this concept is depicted in the image. 
%ML: At the end we present results of another prompt to show that LVLMs are able to respond to framing when necessary.
%ML: Function: Belongs with the description of expected behavior of LVLM. 
%ML Note: No paragraphs of only one sentence
%ML Function: How the behavior that we observe deviates from the expected behavior
% The text-only setting corresponds to a two-step prompting in which recognition and attribute generation are separated, whereas the text+image setting corresponds to a one-step prompting in which attribute generation occurs directly from visual input.
However, our analyses show that LVLMs fall short of this goal or do not consistently achieve it.
%the attribute descriptions of the target concept do not always remain stable across these input conditions.
Specifically, compared with the text-only setting, adding images increases physical-term proportions for animal breeds (from 16.7\% to 32.8\%) and decreases sensory-term proportions for food categories (from 83.6\% to 73.2\%) among the most frequent attribute terms, averaged over all evaluated models and two framings on correctly recognized images.
% For animal breeds, physical terms increase from 18\% in the text-only setting to 45\% for subject-focused images and 40\% for subject-in-situation images. 
Within the text+image setting, attribute-term distributions differ significantly between the two framing types for 41.4\% of animal and 71.4\% of food model--concept pairs.
Analysis of the directional terms shows that physical-term proportions for animals and sensory-term proportions for food are generally higher under subject-focused than subject-in-situation framing.
% These patterns vary across models and concepts.

%We additionally test image-relevant prompts asking about the image itself, using two models on the same animal image set.
For comparison, we include a study of a case where an effect of visual framing on attribute-based descriptions is appropriate: image-relevant prompts.
As expected, differences between framings are more pronounced for image-relevant prompts than for concept-relevant prompts, confirming that the model does have a basic capacity to focus on the concept in the case of concept-relevant prompts.
The larger point of this paper is that this basic capacity is not adequate for robustness, since visual input and its framing should have no measurable effect in the case of concept-relevant prompts.
%generally larger than under concept-relevant prompts and statistically significant for nearly all breeds in both models.
%Such variation may be appropriate when LVLMs are asked to describe an image.
%Together, these results show that LVLMs respond to framing under image-relevant prompts, but do not ignore it under concept-relevant prompts.
%Overall, our findings show that LVLM-generated attribute descriptions of the target concept are affected not only by the recognized target concept, but also by visual input and its framing.
Overall, our findings highlight the importance of considering the impact of visual inputs and how they are framed in research on LVLMs' robustness.

\section{Related Work}
\paragraph{Visual framing and intent.}
Visual framing reflects how a subject is presented, rather than only what is depicted.
~\citet{Rodriguez11framinglevels} distinguish four levels of visual framing: denotative, stylistic-semiotic, connotative, and ideological, covering literal content, visual composition, and broader meanings.
In multimedia retrieval, intentional framing describes photographers' choices in presenting a subject~\citep{riegler14howhow}, while photographer intent supports image diversification beyond topical content~\citep{wang17beyond}.
Studying perceived intent in social media images, Intentonomy shows that object and contextual information contribute differently across intent categories~\citep{jia2021intentonomy}.
These perspectives motivate our focus on subject--context relationships, which we operationalize through subject-focused and subject-in-situation images as an experimental contrast rather than an exhaustive framing taxonomy.
Together, these perspectives highlight the role of subject presentation and context in image interpretation.
Within LVLM research, recent work evaluates whether models can classify predefined visual frames in news imagery, treating framing as an output category to be predicted~\citep{lu26LVLMFraming}.
Instead, we study framing as an influential factor that may affect generated attribute descriptions by comparing subject-focused and subject-in-situation images, rather than predicting framing.

\paragraph{Bias in LVLMs.}
Prior work studies social biases in generated text~\citep{howard2025uncovering}, fairness in open- and closed-ended Visual Question Answering (VQA)~\citep{zhang2024vlbiasbench}, spurious 
%PM maybe use the full meaning of VQA and then use the abbreviation as it is not explained what it means. But this is a suggestion.
visual or textual correlations~\citep{ye2026mm}, prior-knowledge bias in visual tasks~\citep{vo2026vlmsbiased}, and answer-token or position bias in multiple-choice VQA~\citep{asgarov2025selection}.
Our work complements these studies by examining framing-related bias in LVLM-generated attribute-based descriptions of the concept: visual input and its framing can shift which attributes LVLMs emphasize, even when the concept is correctly recognized.

\section{Evaluation Methodology}
\label{sec:method}
We evaluate the LVLM-generated attribute-based descriptions elicited by concept-relevant prompts for each target concept under two input settings (cf. Figure~\ref{fig:illustration}): text-only and text+image settings. 
We further analyze the text+image setting by comparing two visual framing types: subject-focused and subject-in-situation images.

\subsection{Collecting Attribute-based Descriptions}
\label{sec: collect_attr}
Our experiments require a representative set of LVLM responses (response set) for each of the settings, which will allow us to compare the distributions of the attribute terms (adjectives and descriptive phrases) that the LVLM produces across settings. 
We follow the methodology of~\citet{visual_text_style}, which achieves diversity using comparable variants of the prompt and re-prompting to collect a large set of responses for each model studied.

Within the text+image setting, we create sets of subject-focused and subject-in-situation images by manual inspection and selection.
Since the number of available images is limited, we expand these sets by creating multiple versions of each image by applying image-processing operations, which reduces possible dependence on specific pixel patterns. 
To isolate variation in attribute generation from failures in target recognition, for each image version, we confirm that the LVLM can correctly recognize the concept it contains before including it in the set of images used to collect attribute outputs.

\subsection{Comparing Attribute-term Distributions}
For text-only vs.\ text+image comparison, we focus on the frequency with which the LVLM produces attributes in the response sets.
For the LVLM response set of each setting, 
we rank attribute terms by their relative frequency and compare the top-$K$ most frequent terms. 
The analysis focuses on highly frequent terms to maintain robustness to the relatively smaller size of the text-only response set, which, in contrast to the text+image response sets, does not contain responses for image variants.

Within the text+image setting, we compare the distributions of the subject-focused and the subject-in-situation response sets for each model and target concept.
We quantify their difference using Total Variation Distance (TVD):
\[
\mathrm{TVD}(P_{c}^{\mathrm{sf}}, P_{c}^{\mathrm{sis}})
=
\frac{1}{2}\sum_{w \in \mathcal{V}_{c}}
\left|P_{c}^{\mathrm{sf}}(w)-P_{c}^{\mathrm{sis}}(w)\right|,
\]
where $\mathcal{V}_{c}$ is the vocabulary of terms observed across these response sets, $P_{c}^{\mathrm{sf}}$ and $P_{c}^{\mathrm{sis}}$ are the empirical relative-frequency distributions of attribute terms for concept $c$ under the two framing types.
TVD ranges from 0 to 1, with larger values indicating larger distributional differences.

We assess statistical significance using a permutation test for each model and concept.
We randomly shuffle framing labels among images within the concept, keeping all outputs associated with each original image and its processing variants together.
We recompute TVD after each permutation to obtain a null distribution under the assumption that framing labels are exchangeable within the concept.
We use 5,000 permutations.
We calculate the one-sided $p$-value by adding one to the number of permuted TVD values at least as large as the observed value and dividing by 5,001, avoiding zero $p$-values.
A $p$-value smaller than 0.05 indicates that the observed difference between framing types is unlikely under random label assignment.

\subsection{Directional Lexical Analysis }
To interpret the distributional differences between the two framing types, we conduct a directional lexical analysis for each model and concept using the unsigned log-likelihood ratio statistic~\citep{freq_profiling}.
Let $a$ and $b$ denote the counts of an attribute term $w$ in the subject-focused and subject-in-situation response sets, respectively, and let $N_{\mathrm{sf}}$ and $N_{\mathrm{sis}}$ denote their total attribute term counts.
Under the null hypothesis of equal relative frequency, the expected counts are
\[
E_{\mathrm{sf}}=\frac{N_{\mathrm{sf}}(a+b)}{N_{\mathrm{sf}}+N_{\mathrm{sis}}},
\qquad
E_{\mathrm{sis}}=\frac{N_{\mathrm{sis}}(a+b)}{N_{\mathrm{sf}}+N_{\mathrm{sis}}}.
\]
We compute the log-likelihood score as
\[
LL(w)=2\left(
a\log\frac{a}{E_{\mathrm{sf}}}
+b\log\frac{b}{E_{\mathrm{sis}}}
\right),
\]
where zero observed counts contribute zero to the sum.
Larger scores mean a larger deviation from equal relative frequency across the two response sets.
Because the score is unsigned, we determine the direction of association by comparing the term's relative frequencies, $a/N_{\mathrm{sf}}$ and $b/N_{\mathrm{sis}}$.
This normalization accounts for differences in the total number of attribute terms collected under the two framing types.
We assign each term to the framing type in which it has the higher relative frequency, then rank terms within each type by their log-likelihood scores in descending order.
This produces two directional keyness lists for each concept: attribute terms most strongly associated with subject-focused images and terms most strongly associated with subject-in-situation images.

\section{Experiments}
\label{sec:experiments}
\subsection{Experimental Setup}
\paragraph{Models.}
We evaluate seven models: five open-source models, the 3B, 7B, and 32B instruction-tuned variants of Qwen2.5-VL~\citep{qwen2.5-VL}, InternVL3-8B~\citep{zhu2025internvl3}, and LLaVA-NeXT (Llama-3-8B)~\citep{li2024llavanext-strong}, and two closed-source models, GPT-4o-mini~\citep{gpt4o_mini} and Gemini3.5-Flash-Lite~\citep{googledeepmind2026gemini35flashlite}. 
We use the Hugging Face Transformers implementation of InternVL3-8B (OpenGVLab/InternVL3-8B-hf) and access the closed-source models through their respective APIs. 
Throughout the paper, we abbreviate the Qwen variants as Qwen2.5-VL-3B, Qwen2.5-VL-7B, and Qwen2.5-VL-32B, respectively, and refer to the LLaVA-NeXT (Llama-3-8B) model as LLaVA-NeXT-8B.
For Qwen variants, we use \texttt{torch\_dtype=\texttt{torch.bfloat16}}.
We set the temperature to 0 for all queries.
All experiments were run on internal servers with NVIDIA RTX A6000, RTX A5000, and RTX 3090 GPUs, for approximately 1,700 GPU hours in total; the two closed-source models were accessed via APIs and required no local compute.

\paragraph{Image sets.}
We construct two image sets, one for animal breeds and one for food categories, with target concepts from Oxford-IIIT Pet~\citep{oxfordpets} and Food-101~\citep{food101}, respectively.
For animal breeds, we also collect images from Wikimedia Commons via its public API. 

For recognition filtering, we generate five variants of each candidate image using spatial cropping, Non-Local Means denoising, contrast scaling, brightness scaling, and sharpening.
Processing parameters are provided in Appendix~\ref{sec:image_processing_para}.
We then ask each evaluated model to identify the subject of the original image and each processed variant using a direct recognition prompt listed in Appendix~\ref{sec:prompts}.
An original image is retained only if all models correctly recognize its subject in the original and all five variants.
After recognition filtering, we exclude seven breeds with insufficient qualifying images: Abyssinian, Birman, Bombay, Ragdoll, Havanese, Japanese Chin, and English Cocker Spaniel.

We manually annotate the remaining original images by visual framing: subject-focused images present the subject as the dominant visual element, whereas subject-in-situation images present the subject within a broader activity, interaction, or environmental context. 
Annotation details are provided in Appendix~\ref{sec:visual_framing_type_annotation}.

Finally, we construct balanced evaluation sets with equal numbers of subject-focused and subject-in-situation images within each concept.
We prioritize broader concept coverage for animals (30 breeds, 8 images per framing per breed) and more images per category for food (8 categories, 30 images per framing per category), yielding 480 original images in each set. 
The food categories are Edamame, French toast, Guacamole, Hot dog, Nachos, Onion rings, Pizza, and Sushi.

\paragraph{Attribute collection.}
Following~\citet{visual_text_style}, we collect attribute outputs using five prompt variants and five repetitions in both settings.
In the text-only setting, models generate attribute descriptions of the target concept only from the target concept name, yielding 25 outputs per concept.
In the text+image setting, we use eight images per animal breed and 30 per food category for each framing, and include the original image together with five image variants per image, yielding 1,200 and 4,500 outputs per concept per framing, respectively.
Full prompts are provided in Appendix~\ref{sec:prompts}.

\begin{figure*}[t]
    \centering
    \includegraphics[width=\linewidth]{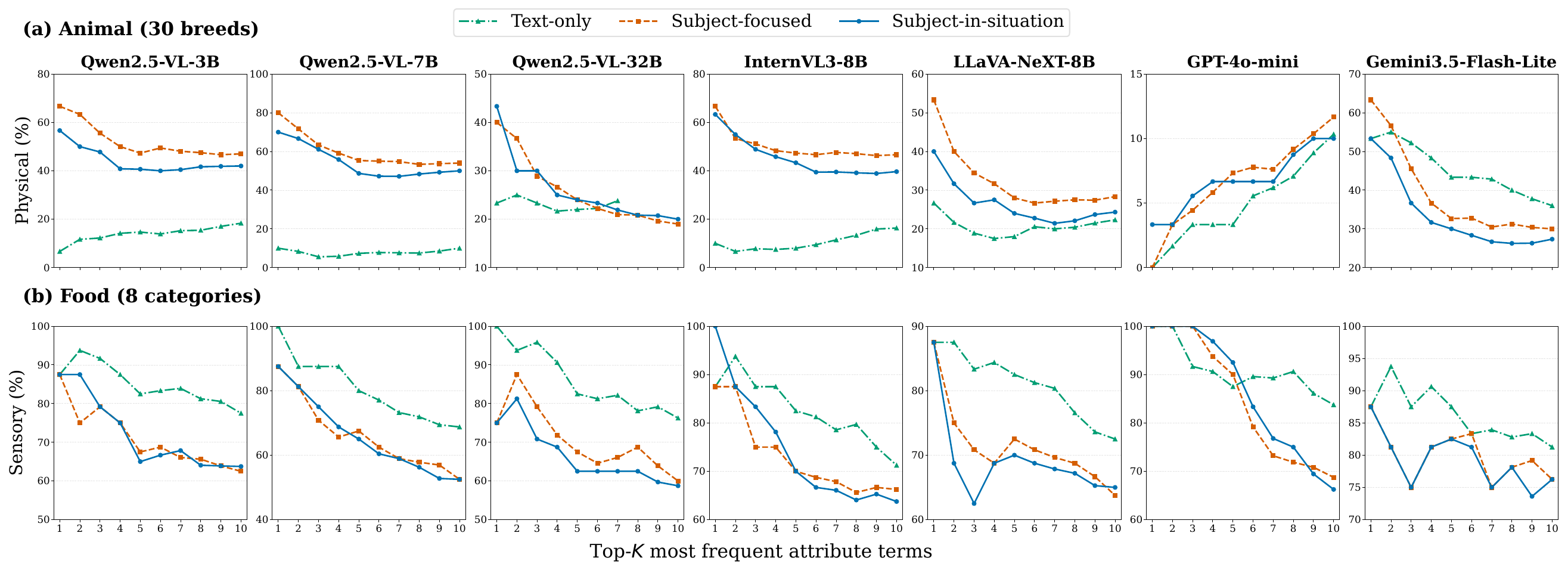}
    \caption{Percentage of physical (animal) and sensory (food) terms among the top-$K$ most frequent attribute terms in the attribute-term distributions under text-only and text+image (subject-focused and subject-in-situation) settings. The upper and lower panels share model names, x-axis tick values, and x-axis labels. Visual input 
    %PM is this explaining the x-axis and y-axis correctly?
    shifts attribute descriptions of the target concept, generally toward more physical terms for animal breeds and fewer sensory terms for food categories.}
    \label{fig:text_image_input_comparison}
\end{figure*}
\begin{figure*}[h]
    \centering
    \includegraphics[width=1\linewidth]{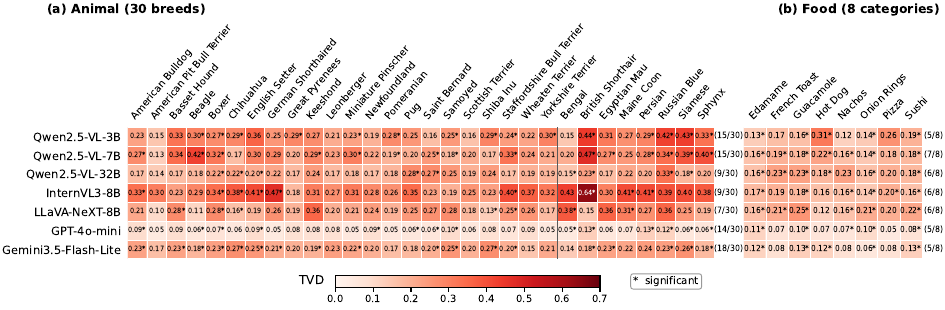}
    \caption{
Per-concept TVD between subject-focused and subject-in-situation attribute-term distributions under concept-relevant prompts.
Rows represent models and columns represent concepts.
Asterisks mark statistically significant differences (image-block permutation test, $p<0.05$).
Row-end counts show significant/total concepts.
Across all models, visual framing significantly affects attribute descriptions of a subset of concepts, even when the concepts are correctly recognized.
}
\label{fig:per_concept_tvd}
\end{figure*}

\paragraph{Attribute term extraction and annotation.}
Following prior work~\citep{visual_text_style}, we use Llama-3.1-8B~\citep{llama3_1_8b} to standardize raw responses and extract attribute terms.
For animal breeds, we categorize extracted terms into physical and non-physical terms.
Physical terms describe visually observable properties, such as morphology, coat, color, size, body structure, and facial features.
Non-physical terms include temperament-related, ambiguous, or evaluative terms that do not unambiguously describe visible physical appearance.
For food categories, we categorize terms as sensory and non-sensory.
Sensory terms describe perceptual characteristics of the food itself, including appearance, taste, texture, smell, and temperature.
Non-sensory terms include ingredient identity and content, health and nutrition, preparation, cultural associations, and how or where the food is served or eaten, as well as ambiguous or evaluative terms that do not specify a sensory property.
Details are provided in Appendix~\ref{sec:attribute_type_annotation}.

\subsection{Experimental Results}
\label{sec:results}
\paragraph{Text-only vs. text+image setting: visual input changes attribute-term-type proportions.}
Figure~\ref{fig:text_image_input_comparison} compares the percentage of physical terms (animal breeds) and sensory terms (food categories) among the top-$K$ most frequent terms in the text-only and text+image response sets.
For animal breeds, the increase with visual input (i.e., text-only vs. subject-focused + subject-in-situation) is clearest for Qwen2.5-VL-3B, Qwen2.5-VL-7B, and InternVL3-8B.
However, Qwen2.5-VL-32B shows similar proportions across text-only input and both framing types at $K=5$--$10$, while Gemini3.5-Flash-Lite shows lower proportions under both framing types than text-only input at $K=3$--$10$.
For food categories, sensory-term proportions are generally lower with image input.
At $K=10$, both framing types yield lower proportions than the text-only setting for all models.
These results demonstrate that adding images impacts the stability of attribute-based descriptions.
% Thus, adding images changes the attribute types emphasized, but the direction depends on model and image set.

\paragraph{Text+image setting: framing affects attribute-term distributions across models.}
Within the text+image setting, we compare attribute-term distributions between subject-focused and subject-in-situation images under concept-relevant prompts.
Figure~\ref{fig:per_concept_tvd} shows statistically significant differences for subsets of concepts in every evaluated model.
Across all model--concept pairs, these differences are significant for 87 of 210 animal comparisons (41.4\%) and 40 of 56 food comparisons (71.4\%) at $p<0.05$.
All seven models show significant differences for the breed Boxer and the food categories Edamame, Guacamole, Onion rings, and Sushi.
In contrast, no model shows a significant difference for American Pit Bull Terrier, Leonberger, Scottish Terrier, or Wheaten Terrier.

\begin{figure}[h]
    \centering
    \includegraphics[width=\linewidth]{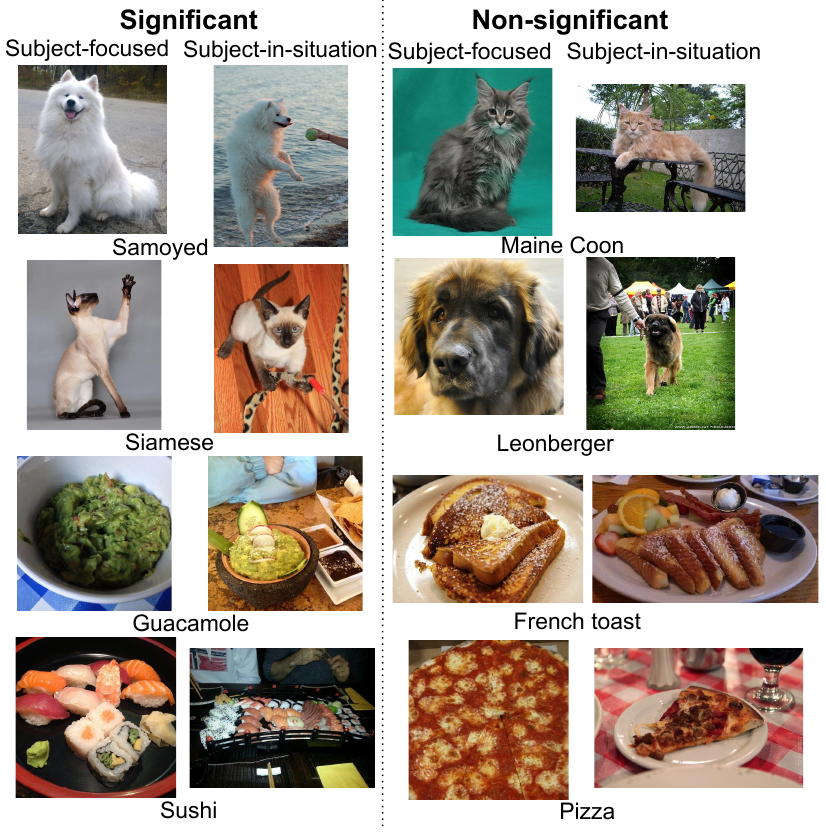}
    \caption{Text+image setting across framing types: significant and non-significant cases for both Qwen2.5-VL-3B and GPT-4o-mini. Significant cases tend to show clearer contrasts in subject presentation and context between two framing types. }
    \label{fig:non-significant_case}
\end{figure}

We illustrate the images from significant and non-significant cases for Qwen2.5-VL-3B and GPT-4o-mini (Figure~\ref{fig:non-significant_case}).
In the illustrated significant animal cases, subject-focused images depict the animal as a centered and visually dominant subject, often with a clear view of the body and a plain or unobtrusive background, whereas subject-in-situation images include activity, interaction, or environmental context.
A similar contrast appears in the significant food examples: subject-focused images provide close views of the food with limited surrounding context, whereas subject-in-situation images include people and a broader table setting.
The non-significant animal examples contain other visual differences, such as partial close-ups or visually heterogeneous subjects within the same breed.
These examples are illustrative; significance is assessed over the full image sets, not from the visual contrast between individual image pairs.

\begin{figure*}[t]
    \centering
    \includegraphics[width=\linewidth]{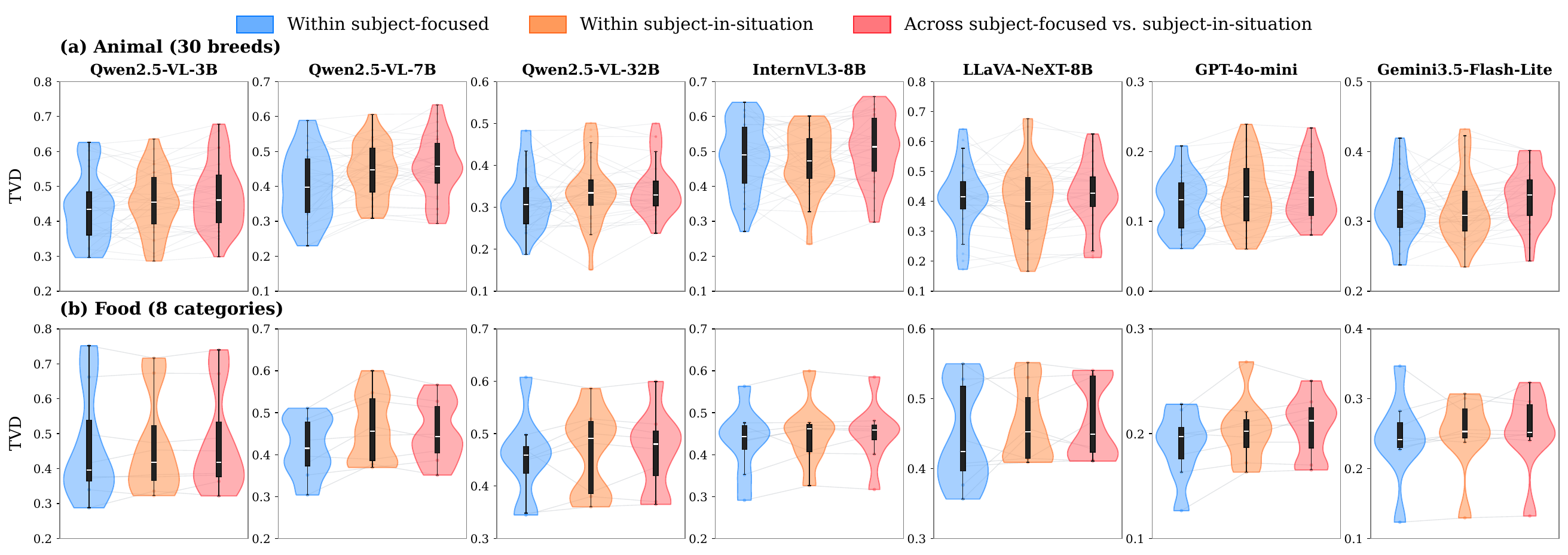}
    \caption{
Image-level pairwise TVD within and across framing types under concept-relevant prompts.
Each point represents the mean pairwise TVD for one concept within subject-focused images, within subject-in-situation images, or across framing types.
Lines connect the three values for the same concept.
The upper and lower panels share model names.
Visual framing can be an important source of variation in attribute descriptions of the concept, but its contribution varies across models and concepts.
}
\label{fig:within_across_tvd}
\end{figure*}

We also compare per-concept mean pairwise TVDs between image-level attribute-term distributions within and across framing types (Figure~\ref{fig:within_across_tvd}).
The distributions of these values overlap, and across-framing means are not consistently higher than the within-framing means for both framings.
Together with Figure~\ref{fig:per_concept_tvd}, this indicates that framing can play an important role in shaping attribute descriptions of the target concept, though its influence differs across models and concepts.

\begin{figure*}
\centering
    \includegraphics[width=1\linewidth]{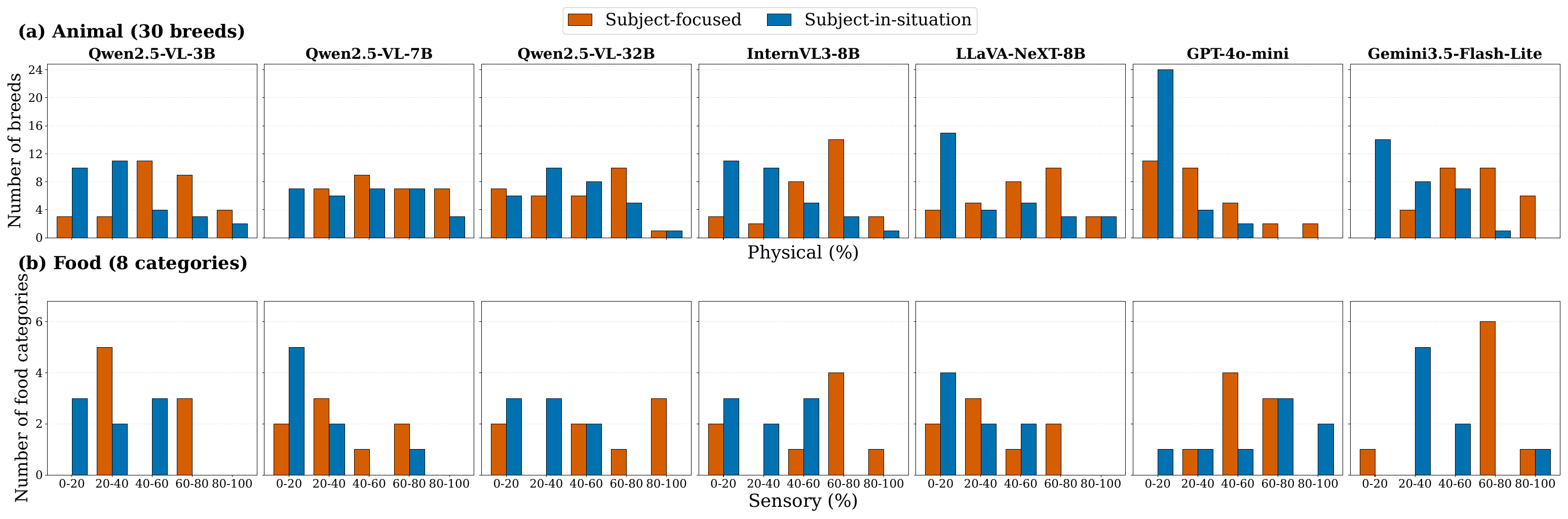}
    \caption{Text+image settings: Percentages of physical (animal) and sensory (food) terms among the top-5 directional terms in the keyness list for each framing type. Bars show the number of concepts in each percentage range. The upper and lower panels share model names and x-axis tick values. Subject-focused images generally yield more physical terms for animal breeds and sensory terms for food categories than subject-in-situation images.}
    \label{fig:percentage_term}
\end{figure*}

\paragraph{Text+image setting: framing affects which attribute-term type models emphasize.}
We next examine which attribute terms characterize the difference between the two framing types.
Figure~\ref{fig:percentage_term} shows physical-term proportions for animal breeds and sensory-term proportions for food categories among the top-5 directional terms from each of the two directional keyness lists.
For animal breeds, all seven models have more breeds in the 60--100\% physical-term range under subject-focused than subject-in-situation framing.
The counts are 13 versus 5 breeds for Qwen2.5-VL-3B
and 17 versus 4 for InternVL3-8B.
For food categories, six of the seven models show the same pattern in the 60--100\% sensory-term range, including 5 versus 0 categories for InternVL3-8B and 7 versus 1 for Gemini3.5-Flash-Lite.
GPT-4o-mini shows the opposite pattern, with 3 versus 5 categories.
Overall, subject-focused images are generally associated with higher physical-term proportions for animal breeds and sensory-term proportions for food categories than subject-in-situation images.

\section{Concept- vs.\ Image-Relevant Prompts}
\label{sec:concept_vs_image}
As noted in the introduction, framing can matter when a prompt asks about the image itself rather than the general properties of the target concept presented by the subject depicted in the image.
We therefore compare concept-relevant and image-relevant prompts on the same images to examine how attribute descriptions vary across framing types under two types of prompts.
Image-relevant prompts request image attributes, which describe properties of the depicted subject, scene, or photograph, including appearance, current state, brightness, and sharpness.

\begin{figure}[t]
    \centering
    \includegraphics[width=\linewidth]{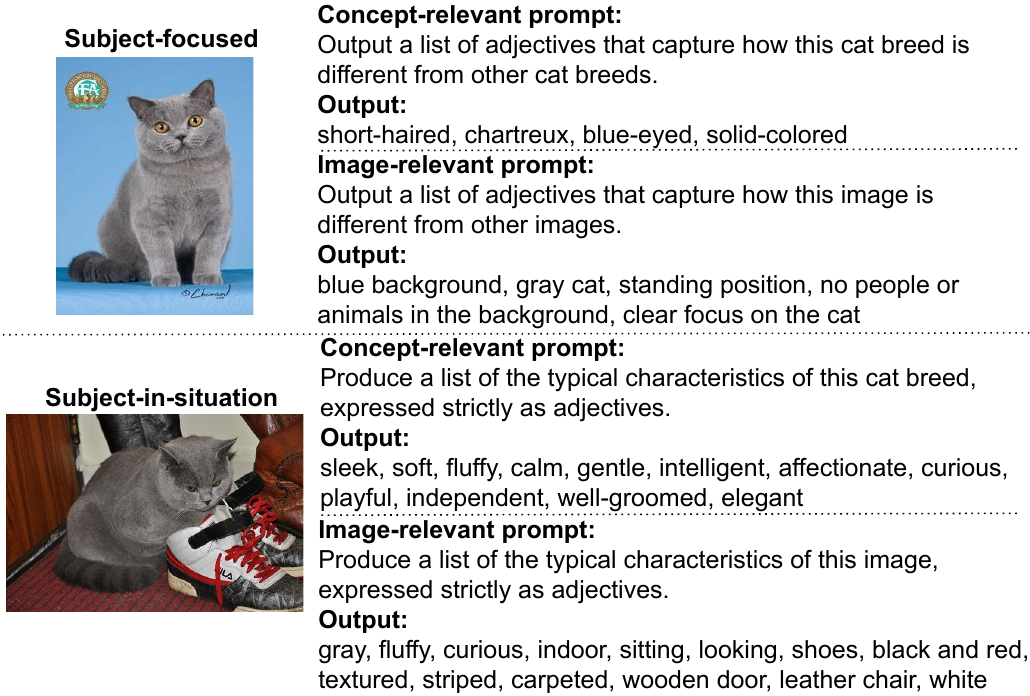}
    \caption{Example Qwen2.5-VL-3B outputs for two British Shorthair images with two framings, each queried with paired concept- and image-relevant prompts.
    Image-relevant prompts shift descriptions toward image-specific details of depicted content.
    }
    \label{fig:image_relevant_examples}
\end{figure}

Figure~\ref{fig:image_relevant_examples} illustrates this comparison using Qwen2.5-VL-3B and two British Shorthair images, one from each framing type.
Concept-relevant prompts request general attributes of the breed, with outputs including physical and non-physical terms such as ``short-haired'', ``calm'', and ``affectionate''.
For the same images, image-relevant prompts also produce image-specific details: ``blue background'' and ``clear focus on the cat'' for the subject-focused image, and ``shoes'', ``carpeted'', and ``wooden door'' for the subject-in-situation image.
These examples confirm that changing the requested description target in the prompt brings details of the depicted subject, scene, and photograph into the response.

We next quantify the differences in attribute-term distributions between framing types under image-relevant prompts.
We evaluate Qwen2.5-VL-3B and GPT-4o-mini on the same animal evaluation image set.
In the five concept-relevant prompt templates, we replace ``this cat/dog breed'' with ``this image'' and, where applicable, ``other cat/dog breeds'' with ``other images''.
All other wording, output requirements, and generation settings remain unchanged.
We standardize responses using Llama-3.1-8B, following the same procedure as in the preceding experiments.
The full prompts are provided in Appendix~\ref{app:image_relevant_prompts}.

Figure~\ref{fig:tvd_image_relevant} reports the TVDs between subject-focused and subject-in-situation attribute-term distributions for each breed.
The differences are statistically significant for 29 of 30 breeds in Qwen2.5-VL-3B and all 30 breeds in GPT-4o-mini.
The TVDs are generally higher than those for the same models and breeds under concept-relevant prompts (Figure~\ref{fig:per_concept_tvd}).
Such differences can be appropriate under the image-relevant prompts because the images from different framing types contain different details to describe.
By contrast, concept-relevant prompts should produce similar attribute-term distributions across framing types (TVD is near 0), since the target concept remains unchanged.
Together with the results under concept-relevant prompts, this shows that models can respond to framing when it is relevant. However, the key message of this paper is that they also remain sensitive to framing when only the concept should matter.

\begin{figure}[h]
    \centering
    \includegraphics[width=\linewidth]{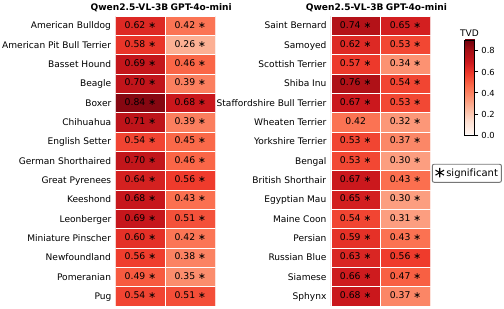}
    \caption{Per-concept TVD between subject-focused and subject-in-situation attribute-term distributions under image-relevant prompts.  Wheaten Terrier for Qwen2.5-VL-3B is the only non-significant case.
    Large framing-related variation is observed under image-relevant prompts, which can appropriately reflect image differences in depicted content.
    }
    \label{fig:tvd_image_relevant}
\end{figure}

\section{Conclusion and Outlook}
\paragraph{Conclusion.}
Across seven evaluated LVLMs and two image sets, we have shown that the presence of visual input and its framing can affect LVLM-generated attribute descriptions, even when the target concept is correctly recognized.
Compared with the text-only setting, adding images often increases physical-term proportions for animal breeds and decreases sensory-term proportions for food categories among the most frequent attribute terms.
Within the text+image setting, significant differences in attribute-term distributions show that concept-relevant prompts do not ensure stable descriptions across framing types.
Among the directional terms associated with framing, subject-focused images are generally associated with higher proportions of physical terms for animal breeds and sensory terms for food categories than subject-in-situation images.
These shifts can, in some contexts, be plausible rather than obviously incorrect, making them challenging to study.
With controlled and systematic evaluation, however, we have shown that visual input and its framing affect LVLM-generated descriptions.

% outlook
% in our experiments, we observe that the model struggle distinguishing the the ownership of attributes (e.g., the image attributes and the subject attributes). 
% It is a potential issue of the models.
\paragraph{Outlook.}
In Figure~\ref{fig:image_relevant_examples}, the image-relevant response for the subject-in-situation image still begins with attributes of the cat, suggesting that the model may generate reasonable attributes without clearly understanding what the prompt asks it to describe.
% However, a description of the whole image can include the cat's visible attributes, so mentioning these attributes alone does not show that the model misunderstood the prompt.
Recent studies have shown that VLMs struggle to link attributes to the correct objects~\citep{yuksekgonul2023when}, and in language models, models may not fully use sentence structure~\citep{sinha-etal-2021-unnatural} or prompt meaning~\citep{webson2022prompt}, even when they perform the task successfully.
Future work could test whether generated attributes match the target specified by the prompt and how this varies across framing types.

%Future research should investigate the affect of visual framing on LVLMs within complete pipelines, such as perception modules for autonomous agents mentioned in Sec.~\ref{sec:intro}. 
Formulating concept-relevant prompts in two steps, concept identification followed by concept characterization, will stabilize the unwanted impact of framing in cases in which the first step is carried out correctly, reducing the second step to the text-only case we have evaluated here.
However, future work should investigate complete pipelines, such as perception modules for autonomous agents mentioned in Sec.~\ref{sec:intro}, to ensure that such mitigation is used always and only where needed and does not create an error cascade when the LVLM fails in concept identification.

%text+image setting use a one-step prompting strategy for generating attribute descriptions directly from the image and the concept-relevant prompt in a single call.

%A practical approach for mitigating framing effects under one-step prompting is two-step prompting, which separates concept recognition from attribute generation.
%Specifically, the model would first identify the target concept in the image, then generate its attributes from the predicted concept name in a new text-only call without the image or prior conversation.
%Our existing text-only setting corresponds only to the attribute-generation stage with a correct concept name, rather than an evaluation of the complete workflow.
%Future work could test whether the full two-step workflow improves the stability of attribute descriptions.
% Future work should test whether similar effects appear in open-ended outputs such as captions or assistive explanations.

\section*{Limitations}
\paragraph{Domain scope.}
Our study uses animal breeds from the Oxford-IIIT Pet dataset and food categories from the Food-101 dataset.
While these domains provide controlled and manually verifiable test sets, they may not fully capture framing effects in other visual domains.
Future work should test for effects arising in specific applications such as image captioning, assistive technologies, and perception modules of intelligent agents, extending the settings studied to vehicles, scenes, places, objects, or activities.

\paragraph{Visual framing granularity.}
We consider two high-level visual framing types: subject-focused and subject-in-situation.
This binary contrast is useful for isolating a clear framing difference. However, visual framing can involve distinctions at different resolutions. For example, images of a rose in a botanical photo and a rose in the middle of a large rose garden differ in terms of framing with a large visual distinction.
In contrast, a portrait of a person and a mugshot are also different in terms of framing, but with a visual distinction that is readily evident to a person looking at the image, yet represents a more subtle visual distinction.
Future work should extend the experiments to include other framings.

\paragraph{Attribute term type annotation.}
Our attribute type annotation is intentionally coarse.
We use binary distinctions between physical and non-physical terms for animal breeds and between sensory and non-sensory terms for food categories to capture the main semantic direction of the observed framing effect.
This grouping simplifies the space of possible attributes: it allows us to test whether visual framing shifts outputs toward physical/sensory descriptions, but it does not provide a fine-grained taxonomy of attributes.
It should also not be interpreted as a claim that model outputs constitute measurements of breed morphology or temperament in the ethological sense.
Similarly, generated attributes for food categories should not be interpreted as validated assessments of the food's sensory attributes, ingredient content, or nutritional value.
Future work should use more detailed categories to distinguish attribute terms.

\section*{Ethical Considerations}
This work analyzes how visual input and visual framing affect attribute-based descriptions generated by LVLMs. 
Our experiments use natural images of animal breeds and food categories, and do not involve human subjects, private personal information, or sensitive demographic attributes. 
The generated attributes are analyzed only to study model behavior.

The main ethical implication is that LVLM-generated descriptions may be associated with the target concept, but can still be affected by visual framing after correct recognition. 
In real-world applications such as assistive technologies, image retrieval, content moderation, or autonomous perception, users may over-interpret such outputs as stable properties of the recognized concept. 
This issue could become more consequential for images involving people, social groups, occupations, or culturally sensitive categories, where framing-induced shifts may reinforce stereotypes or produce misleading descriptions.

Therefore, our findings highlight the importance of evaluating not only recognition accuracy, but also the stability of downstream semantic descriptions under different image presentations. 
Users should be cautious when using LVLM-generated attributes as knowledge of the target concept, as these attributes can vary with how the concept is visually presented.

% \section*{Acknowledgments}

\bibliography{reference}

\clearpage
\appendix
\section{Experimental Details}
\subsection{Image Processing Operations}
\label{sec:image_processing_para}
We use a crop ratio of $0.8$, denoising with \texttt{cv2.fastNlMeansDenoisingColored} and filtering strength $h=5$, contrast and brightness factors of $1.05$, and unsharp masking with \texttt{unsharp\_percent}$=70$.

\subsection{Prompt Templates}
\label{sec:prompts}
This appendix lists the prompts used for recognition filtering and attribute generation. 

\paragraph{Recognition prompt.}
For recognition filtering, we use the following prompt for animal breeds:
\begin{quote}
\emph{Identify the breed of the cat pictured in the image. Answer with the breed name directly.}
\end{quote}
For dog breeds, \emph{cat} is replaced with \emph{dog}.
For food categories, the prompt is:
\begin{quote}
    \emph{
    Identify the category of the food pictured in the image. Answer with the food name directly.
    }
\end{quote}

\paragraph{Concept-relevant attribute generation prompts.}
We use the same five prompt templates in the text-only and text+image settings, with only the reference to the target concept changed across settings.
For animal breeds, the prompt in the text+image setting refers to the breed shown in the image as \emph{this cat breed}.
In the text-only setting, this phrase is replaced by the explicit breed name, e.g., \emph{the cat breed Birman}.
For dog breeds, \emph{cat} is replaced by \emph{dog}.

The five prompt templates are:
\begin{itemize}
    \item \emph{Output a list of the typical attributes of [TARGET], expressed strictly as adjectives.}
    \item \emph{Output a list of attributes that distinguish [TARGET] from other cat breeds, expressed strictly as adjectives.}
    \item \emph{Output a list of adjectives that describe [TARGET].}
    \item \emph{Output a list of adjectives that capture how [TARGET] is different from other cat breeds.}
    \item \emph{Produce a list of the typical characteristics of [TARGET], expressed strictly as adjectives.}
\end{itemize}

Here, [TARGET] is instantiated as \emph{this cat breed} in the text+image setting and as the explicit breed name in the text-only setting, e.g., \emph{the cat breed British Shorthair}.
For example, the first template becomes:
\begin{quote}
\emph{Output a list of the typical attributes of this cat breed, expressed strictly as adjectives.}
\end{quote}
in the text+image setting, and:
\begin{quote}
\emph{Output a list of the typical attributes of the cat breed British Shorthair, expressed strictly as adjectives.}
\end{quote}
in the text-only setting.

For food categories, [TARGET] is instantiated as
\emph{this food category} in the text+image setting and as the explicit category name in the text-only setting, e.g., \emph{the food category guacamole}.
The five prompt templates are:
\begin{itemize}
    \item Output a list of the typical attributes of [TARGET], expressed strictly as adjectives.

    \item Output a list of attributes that distinguish [TARGET] from other food categories, expressed strictly as adjectives.

    \item Output a list of adjectives that describe [TARGET].

    \item Output a list of adjectives that capture how [TARGET] is different from other food categories.

    \item Produce a list of the typical characteristics of [TARGET], expressed strictly as adjectives.
\end{itemize}

\paragraph{Image-relevant attribute generation prompts for animal breeds.}
\label{app:image_relevant_prompts}
We adapt the five templates above by replacing [TARGET] with \emph{this image} and, where applicable, \emph{other cat/dog breeds} with \emph{other images}:
\begin{itemize}
    \item \emph{Output a list of the typical attributes of this image, expressed strictly as adjectives.}
    \item \emph{Output a list of attributes that distinguish this image from other images, expressed strictly as adjectives.}
    \item \emph{Output a list of adjectives that describe this image.}
    \item \emph{Output a list of adjectives that capture how this image is different from other images.}
    \item \emph{Produce a list of the typical characteristics of this image, expressed strictly as adjectives.}
\end{itemize}

\section{Visual Framing Type Annotation}
\label{sec:visual_framing_type_annotation}
\begin{figure*}
    \centering
    \includegraphics[width=\linewidth]{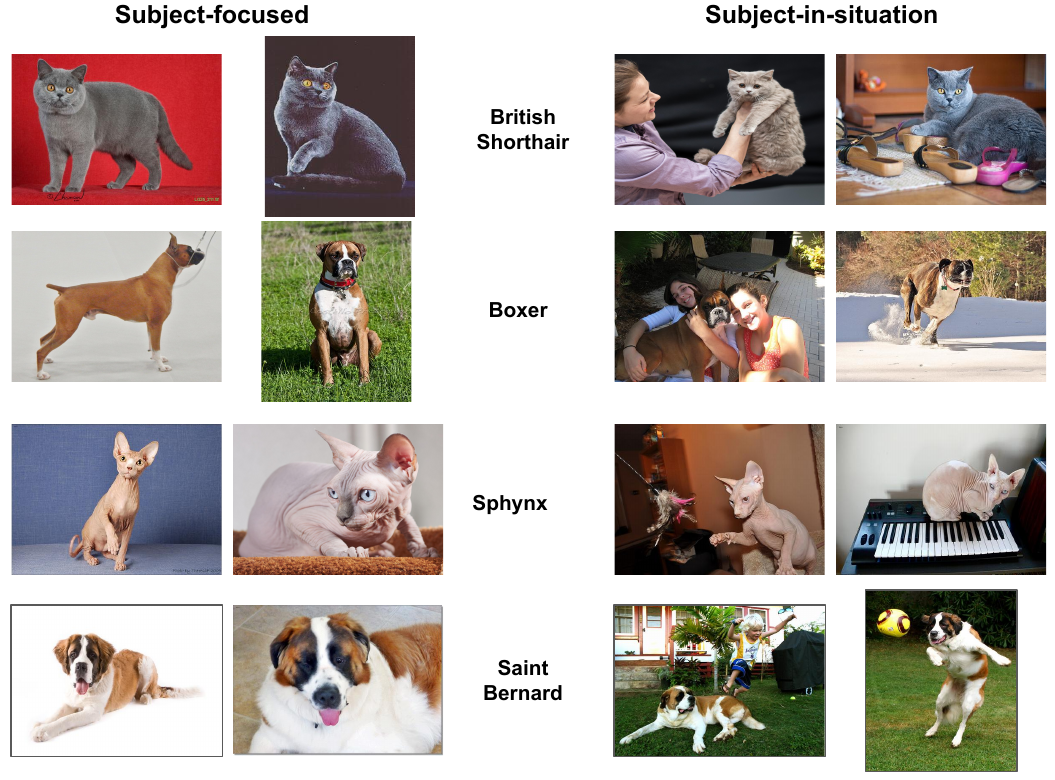}
    \caption{Examples of two visual framing types for animal breeds. Subject-focused images emphasize the target animal itself, while subject-in-situation images present the target animal within an activity, interaction, or environment.
}
    \label{fig:framing_examples_animal}
\end{figure*}

\begin{figure*}[h]
    \centering
    \includegraphics[width=\linewidth]{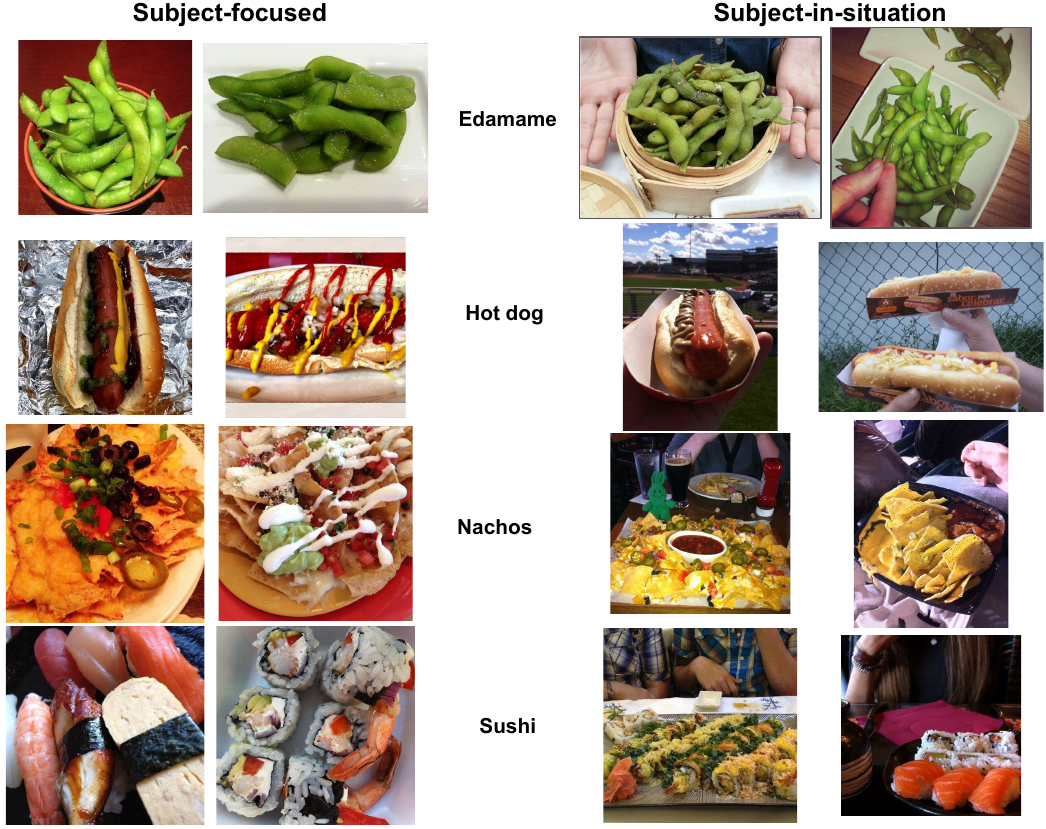}
    \caption{Examples of two visual framing types for food categories. Subject-focused images emphasize the target food itself, while subject-in-situation images present the target food within a context.
}
    \label{fig:framing_examples_food}
\end{figure*}

We annotate images into two visual framing types: subject-focused and subject-in-situation, which operationalizes a controlled contrast in how the same target concept is visually presented.
They are not intended to exhaust all possible types of visual framing.
Instead, they allow us to test whether LVLM-generated attribute descriptions change when the recognized concepts remain the same, but their framing differs.
More examples of the two visual framing types are provided in Figure~\ref{fig:framing_examples_animal} and Figure~\ref{fig:framing_examples_food}.

\paragraph{General principle.}
The annotation concerns how the target concept is visually presented in the image, rather than what the target concept is.
In general, a subject-focused image presents the target subject as the primary object of visual inspection, whereas a subject-in-situation image presents the target subject as part of a broader situation, activity, interaction, or environment.
The distinction is therefore based on two criteria:
(i) the visual dominance of the target subject, and
(ii) the extent to which surrounding contextual elements contribute to the interpretation of the image.

% In our dataset, the target concepts are animal breed concepts.
% All candidate images depict a retained breed concept and pass the recognition filtering described in Section~\ref{sec:recognition_filtering}.
% We then assign a framing type according to how the target animal is presented in the image.
% Images are selected only when the framing type is clear.
% Ambiguous cases are excluded.

\paragraph{Subject-focused images.}
In general, an image is labeled as subject-focused when the target subject is the dominant visual element and the surrounding context is minimal, secondary, or not essential to interpreting the image.
The image primarily presents inspection of the subject itself, rather than interpretation of an event or situation involving the subject.

For animal breeds, a subject-focused image presents the target animal prominently, typically occupying a large portion of the image or appearing centrally.
Such images make visible properties of the animal salient, including morphology, body shape, coat, color, size, posture, facial features, and other visually observable characteristics.
The background may be present, but it should not provide a salient activity, interaction, or environmental situation.
Typical subject-focused examples include close-up or medium-shot photographs of the animal, portrait-like images, studio-like images, images with a clean and context-free background, and images where the animal is standing, sitting, or looking at the camera without a salient ongoing activity.

For food categories, a subject-focused image presents the target food as the dominant visual subject, while the surrounding context is minimal or secondary.
The image primarily supports inspection of the food itself, including its color, shape, and visible surface details.
Typical examples include close views of a dish or a single serving with limited surrounding context.
Plates, bowls, garnishes, or utensils may be present, but they do not dominate the food.
%PM does this sentence belong here? Maybe in the next definition?

\paragraph{Subject-in-situation images.}
In general, an image is labeled as subject-in-situation when the target subject is embedded in a broader visual situation.
The subject remains identifiable, but the surrounding scene, activity, interaction, or environment contributes substantially to the interpretation of the image.
The image provides an interpretation of the subject as participating in, responding to, or being situated within an event or context.

For animal breeds, a subject-in-situation image shows the target animal in an activity, interaction, or environment.
The depicted animal may be playing, running, walking, interacting with a human or another animal, participating in work or sport, exploring an environment, or appearing in a natural or domestic setting where the background is visually meaningful.
Compared with subject-focused images, these images provide stronger contextual cues about behavior, activity, social interaction, or environmental response.
Typical subject-in-situation examples include animals walking or running outdoors, playing with toys, interacting with people or other animals, participating in work or sport, exploring an environment, or appearing in a scene where the background and action are visually meaningful.

For food categories, a subject-in-situation image presents the target food within a broader serving or eating context.
Surrounding elements, such as accompanying dishes, table settings, diners, or hands interacting with the food, contribute to the interpretation of the image.
Typical examples include food being served, held, shared, or eaten, and food shown as part of a wider meal or dining scene.
The target food may remain visually dominant.
Therefore, the distinction concerns the role of the surrounding context, rather than simply its presence or absence.

% \paragraph{Exclusion criteria.}
% We exclude images when the framing type is unclear or mixed.
% Examples include images where the target animal is visually dominant but also engaged in a highly salient activity, images where the background is complex but does not provide interpretable situational context, images with multiple animals where the target animal is unclear, images with heavy occlusion, low resolution, or unusual cropping, and images where the framing could reasonably be assigned to both categories.
% We also exclude images that fail recognition filtering for either evaluated model or under the selected image-processing variations.

\section{Attribute Term Type Annotation}
\label{sec:attribute_type_annotation}
We annotate extracted attribute terms using binary distinctions to interpret the semantic content of LVLM-generated descriptions: \textit{physical} versus \textit{non-physical} for animal breeds, and \textit{sensory} versus \textit{non-sensory} for food categories.
%PM maybe say animal and food domain to clarify it a bit neater?

% It should not be interpreted as a claim that model outputs constitute measurements of breed temperament in the ethological sense.
% Rather, the goal is to identify whether the generated terms describe expected behavioral tendencies of a breed or instead describe non-behavioral properties such as visible appearance or general evaluative impressions.
% This distinction is designed to interpret whether LVLM-generated attribute descriptions emphasize visually observable properties of the subject or instead describe non-visual, behavioral, relational, or evaluative properties.

\paragraph{Physical attribute terms.}
Physical attribute terms describe visually observable properties of the animal, including morphology, coat, color, size, body structure, facial features, or other bodily characteristics.
These terms refer to properties that can, in principle, be inferred from the animal's visible appearance in an image.
Typical examples include:
\begin{quote}
\emph{short-haired}, \emph{long-haired}, \emph{fluffy}, \emph{muscular}, \emph{compact}, \emph{stocky}, \emph{broad-chested}, \emph{small}, \emph{large}, \emph{short-legged}, \emph{long-eared}, \emph{round-faced}, \emph{droopy-faced},  \emph{brindle}, \emph{blue-coated}, \emph{yellow-eyed}, \emph{spotted}, \emph{sleek}, \emph{fox-like}.
\end{quote}
For example, \emph{short-haired} and \emph{long-haired} describe coat properties; \emph{muscular}, \emph{compact}, and \emph{stocky} describe body structure; \emph{long-eared}, \emph{round-faced}, and \emph{droopy-faced} describe visible facial features.

\paragraph{Non-physical attribute terms.}
Non-physical attribute terms do not directly describe visible bodily properties of the animal.
This category includes two main subtypes: temperament-related terms and ambiguous or evaluative terms.

Temperament-related terms describe expected behavioral responses, dispositions, sociability, activity levels, or interaction styles of a breed.
This operationalization is informed by work that ties temperament to behavioral responses in context~\citep{mackay2015consistent}, recent discussion of terminology in animal individuality research~\citep{travnik2026personality}, the AKC Breed Temperament Guide~\footnote{https://www.akc.org/akctemptest/breed-temperament-guide/}, and a cat breed information website~\footnote{https://cats.com/} as practical references.
Because our analysis concerns breed-level model descriptions rather than individual animal behavior tests, we use this literature only to motivate a practical annotation category: terms are treated as non-physical when they describe how the breed is expected to behave, react, interact, or respond, rather than how it visibly looks.

Typical temperament-related terms in the non-physical attribute term type include:
\begin{quote}
\emph{friendly}, \emph{loyal}, \emph{affectionate}, \emph{gentle}, \emph{calm}, \emph{playful}, \emph{curious}, \emph{energetic}, \emph{active}, \emph{social}, \emph{independent}, \emph{protective}, \emph{alert}, \emph{adaptable}, \emph{reserved}, \emph{bold}, \emph{vocal}, \emph{stubborn}.
\end{quote}
These terms describe behavioral orientation rather than visible bodily properties.
For example, \emph{friendly} and \emph{social} describe interaction style; \emph{curious} and \emph{alert} describe response to stimuli; \emph{active}, \emph{energetic}, and \emph{playful} describe activity level or behavioral style.

Ambiguous or evaluative terms are also included in non-physical terms when they do not clearly describe physical appearance.
Examples include:
\begin{quote}
\emph{cute}, \emph{adorable}, \emph{elegant}, \emph{majestic}, \emph{vibrant},  \emph{domesticated}.
\end{quote}
These terms may reflect an overall impression, aesthetic judgment, category status, or mixed interpretation.
For example, \emph{elegant} may refer to movement style or appearance, and \emph{vibrant} may refer to color, liveliness, and general impression.
Because such terms do not unambiguously describe visible bodily properties, we group them with non-physical terms in the binary analysis.

\paragraph{Sensory attribute terms.}
Sensory attribute terms describe perceptual characteristics of the food itself, including its appearance, taste, texture, smell, and temperature~\citep{papies2013tempting}.
These terms refer to properties that can, in principle, be
experienced through the senses, rather than to the food's
nutritional value, cultural associations, or serving and eating context.
Typical examples include:
\begin{quote}
\emph{golden}, \emph{green}, \emph{colorful}, \emph{round},
\emph{ring-shaped}, \emph{savory}, \emph{sweet}, \emph{salty},
\emph{spicy}, \emph{tangy}, \emph{nutty}, \emph{crispy},
\emph{crunchy}, \emph{creamy}, \emph{smooth}, \emph{chewy},
\emph{fluffy}, \emph{juicy}, \emph{tender}, \emph{aromatic},
\emph{smoky}, \emph{warm}, \emph{hot}, \emph{cold}, and
\emph{chilled}.
\end{quote}
For example, \emph{golden}, \emph{green}, \emph{colorful},
\emph{round}, and \emph{ring-shaped} describe appearance;
\emph{savory}, \emph{sweet}, \emph{salty}, \emph{spicy},
\emph{tangy}, and \emph{nutty} describe taste;
\emph{crispy}, \emph{crunchy}, \emph{creamy}, \emph{smooth}, \emph{chewy}, \emph{fluffy}, \emph{juicy}, and \emph{tender} describe texture or mouthfeel;
\emph{aromatic} and \emph{smoky} describe smell or aroma;
and \emph{warm}, \emph{hot}, \emph{cold}, and \emph{chilled} describe temperature.

The annotation concerns the meaning of the complete attribute term, not whether the described property can be verified from an image.
For example, \emph{sweet} and \emph{aromatic} are treated as sensory terms even when they are inferred by the model rather than directly supported by visual evidence.

\paragraph{Non-sensory attribute terms.}
Non-sensory attribute terms do not directly describe perceptual characteristics of the food itself.
This category includes ingredient identity and content, health and nutritional properties, preparation methods, cultural associations, serving practices, and the social or environmental context in which the food is eaten.
Typical examples include:
\begin{quote}
\emph{nutritious}, \emph{healthy}, \emph{protein-rich},
\emph{vegetarian}, \emph{plant-based}, \emph{high in fiber},
\emph{low in calories}, \emph{fried}, \emph{baked},
\emph{grilled}, \emph{steamed}, \emph{battered},
\emph{deep-fried}, \emph{traditional}, \emph{American},
\emph{Japanese}, \emph{artisanal}, \emph{breakfast},
\emph{casual}, \emph{portable}, \emph{handheld},
\emph{shareable}, and \emph{garnished}.
\end{quote}
For example, \emph{nutritious}, \emph{healthy},
\emph{protein-rich}, \emph{high in fiber}, and
\emph{low in calories} describe health or nutritional
properties;
\emph{fried}, \emph{baked}, \emph{grilled}, \emph{steamed},
\emph{battered}, and \emph{deep-fried} describe preparation methods;
and \emph{traditional}, \emph{American}, \emph{Japanese}, and \emph{artisanal} describe cultural associations or conventions.
Terms such as \emph{breakfast} and \emph{casual} describe eating situations, while \emph{portable}, \emph{handheld}, and \emph{shareable} describe how the food may be served or eaten.

Broadly evaluative terms are also included in the non-sensory category when they do not identify a specific perceptual characteristic.
Examples include:
\begin{quote}
\emph{hearty}, \emph{tasty}, \emph{appetizing},
\emph{delicious}, \emph{comforting}, \emph{satisfying},
\emph{indulgent}, \emph{decadent}, \emph{balanced}, and
\emph{refined}.
\end{quote}
These terms express an overall impression, desirability,
satiating quality, or perceived quality without specifying a particular appearance, taste, texture, smell, or temperature.
When a term does not have a clear sensory interpretation, it is therefore grouped with non-sensory terms in the binary analysis.

% \section{Recognition Performance}
% \label{sec:recognition_performance}
% Recall that our study uses only images for which the LVLM is able to correctly identify the target subject.
% The accuracies on the Oxford-IIIT Pet dataset for Qwen2.5-VL-3B-Instruct and GPT-4o-mini models are 0.664 and 0.7683, respectively.

% \begin{figure*}
%     \centering
%     \includegraphics[width=\linewidth]{figures/qwen_per_concept_permutation_null_distributions.png}
%     \caption{Per-concept permutation tests on Qwen2.5-VL-3B-Instruct. For each concept, the histogram shows the null distribution of TVD values obtained by randomly permuting framing labels between subject-focused and subject-in-situation images. The dashed line indicates the observed TVD, and the p-value gives the proportion of permutations with TVD greater than or equal to the observed value. 
%     % Right-tail observations indicate stronger evidence of framing-induced shifts in attribute distributions.
%     }
%     \label{fig:qwen_permutation_test}
% \end{figure*}
% \begin{figure*}
%     \centering
%     \includegraphics[width=\linewidth]{figures/gpt_per_concept_permutation_null_distributions.png}
%     \caption{Per-concept permutation tests on GPT-4o-mini. For each concept, the histogram shows the null distribution of TVD values obtained by randomly permuting framing labels between subject-focused and subject-in-situation images. The dashed line indicates the observed TVD, and the p-value gives the proportion of permutations with TVD greater than or equal to the observed value. }
%     \label{fig:gpt_permutation_test}
% \end{figure*}

\end{document}